\documentclass[letterpaper, 10pt, conference]{ieeeconf}
\IEEEoverridecommandlockouts
\IEEEoverridecommandlockouts

\usepackage{cite}
\usepackage{mathtools}
\usepackage{amssymb,amsfonts}
\usepackage{bbm}
\usepackage{algorithm}
\usepackage{algorithmic}
\usepackage{graphicx}
\usepackage{textcomp}
\usepackage{xcolor}
\usepackage{booktabs}
\usepackage{hyperref}
\usepackage{url}
\usepackage{placeins}

\begin{document}

\title{Hierarchical Reinforcement Learning for Collision-Free
Locomotion of an Underactuated Biped}

\author{
  Jagannath Prasad Sahoo$^{1*}$, Saurabh Kumar$^{1}$,
  Surya Prakash S.K.$^{2}$, Samiran Datta$^{1}$, Abhay Dwivedi$^{1}$ and Amit Shukla$^{1}$%
  \thanks{$^{1}$CAIR, Indian Institute of Technology Mandi, Mandi, India
          \texttt{\{S23118,s25059,s24037\}@students.iitmandi.ac.in}, \texttt{abhaydwivedi10122005@gmail.com}
          and  \texttt{amitshukla@iitmandi.ac.in}}%
  \thanks{$^{2}$SMME, Indian Institute of Technology Mandi, Mandi, India
          \texttt{\{d21077\}@students.iitmandi.ac.in}}%
  \thanks{Video Link:
          \url{https://drive.google.com/file/d/1efTTfWlWawnvGmUdufkGtoBgXsjYLPRd/view?usp=sharing}}
}

\maketitle


\begin{abstract}
A bipedal robot cannot deviate from its path to avoid an obstacle
without disturbing its balance, and this coupling is most severe on
underactuated platforms such as the biped considered here, which has
four actuated joints per leg and no hip or ankle roll. This paper presents
a Hierarchical Reinforcement Learning (HRL) framework in which a
High-Level (HL) policy observes the robot pose, 36 raycast proximity
measurements, moving-obstacle states, and a receding-horizon local
goal, and outputs a body-velocity command
$(v_x, v_y, \omega_{\text{yaw}})$ every ten control steps, while a
velocity-conditioned Low-Level (LL) policy tracks each command
through PD-controlled joint targets. Both policies are trained
jointly with Soft Actor-Critic (SAC). Because the converged gait is
task-agnostic, it is frozen and driven by classical planners over the
same command interface, yielding three controlled baselines,
SAC+A$^\star$, SAC+RRT$^\star$, and SAC+APF. Across 100 evaluation
trials per method in randomized PyBullet environments, the proposed
method reaches the goal in 98.0\% of static and 88.0\% of dynamic
trials, against at most 78.0\% and 68.0\% for the planner hybrids,
with path lengths within 4\% of the A$^\star$ reference, and
ablations confirm that each observation channel and reward term
contributes materially to this performance.
\textit{This present project information and videos are available at} \href{https://jaggu2606.github.io/biped-hrl-locomotion}
{HRL-Biped}.
\end{abstract}


\section{Introduction}

Obstacle-aware walking remains an open problem for three compounding
reasons. A biped is underactuated, so its floating base cannot be
commanded directly and every avoidance maneuver perturbs balance.
Its velocity set is gait-limited and asymmetric, since a biped
needs several steps to change speed or heading, and on our platform
the absence of roll joints further restricts lateral motion. Moving
obstacles demand anticipation, since by the time proximity alone
signals danger a walking robot may no longer have time to step
aside.

\begin{figure}[t]
\centering
\includegraphics[width=1\columnwidth]{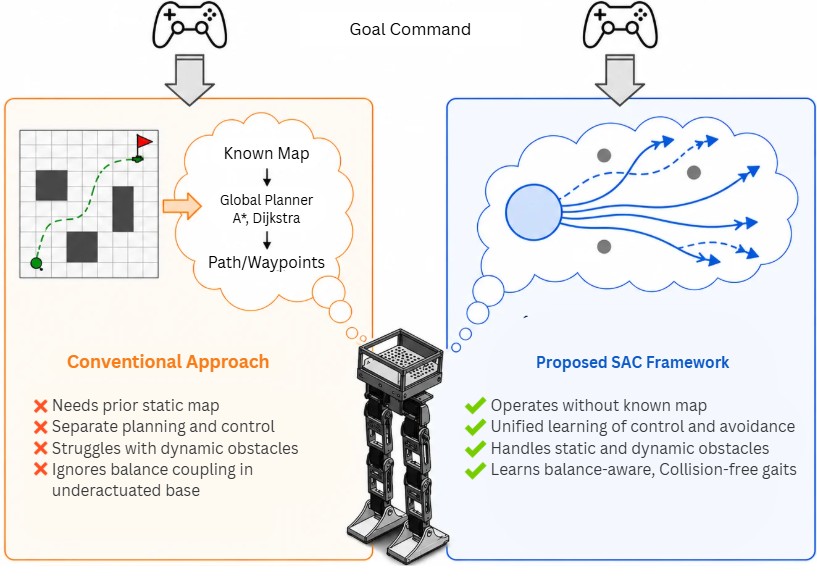}
\caption{Control paradigms for obstacle-aware bipedal locomotion.
Flat RL entangles navigation with gait, decoupled planning ignores
balance coupling, and the proposed hierarchy learns a
velocity-command policy over a learned velocity-conditioned gait.}
\label{fig:intro}
\end{figure}

Existing approaches occupy two extremes (Fig.~\ref{fig:intro}).
Flat end-to-end RL policies map observations directly to joint
targets~\cite{hamze2024}, train slowly, generalize poorly to unseen
layouts, and must relearn locomotion whenever the task changes.
Decoupled pipelines instead track a geometrically planned path with
an independent gait controller~\cite{hart1968,karaman2011,khatib1986},
ignoring balance coupling and, for global planners, moving
obstacles. Neither extreme addresses all three difficulties at
once, which motivates separating navigation from gait generation
without severing the feedback between them.

Given an underactuated biped with eight actuated leg joints, a
known static occupancy map, and moving obstacles whose trajectories
are unknown in advance, the controller must reach a goal at least
6.0~m away without collision or fall, sensing static geometry
through proximity measurements and moving obstacles through their
relative position and velocity, while generating joint-level
actions at 50~Hz.

We solve this with a two-level hierarchy in which both levels are
learned and trained jointly. A lightweight planning layer computes
a global path on the known static map and extracts a
receding-horizon local goal. The high-level (HL) policy decides
where to go next, emitting a body-velocity command
$\mathbf{u}=(v_x, v_y, \omega_{\text{yaw}})$ from the robot pose,
raycast distances, moving-obstacle states, and the local goal. The
low-level (LL) policy decides how to walk, converting each command
into leg-joint targets and executing it for $H=10$ consecutive
control steps before the HL policy is queried again. This
abstraction shortens the effective navigation horizon tenfold, and
joint training lets the gait adapt to the command distribution the
navigation layer actually issues. Because the converged LL policy
is a pure velocity-conditioned gait skill, it can be frozen and
driven by classical planners over the same interface, which we
exploit to build three controlled baselines, SAC+A$^\star$,
SAC+RRT$^\star$, and SAC+APF, in which only the command generator
differs.

The contributions of this work are as follows.
\begin{itemize}
\item A jointly trained two-level HRL framework in which an HL SAC
policy outputs body-velocity commands from pose, raycast,
dynamic-obstacle, and local-goal observations, and a
velocity-conditioned LL SAC policy executes each command for ten
control steps.
\item An HL observation and reward design covering both static
layouts and moving obstacles, including a time-to-collision (TTC)
term inducing anticipatory avoidance.
\item A controlled benchmark against SAC+A$^\star$,
SAC+RRT$^\star$, and SAC+APF hybrids driving the identical frozen
gait, isolating the navigation layer's contribution.
\item An ablation over HL observation channels and reward
components quantifying their effect on success, collision, and
failure rates.
\end{itemize}

\section{Related Work}

Obstacle avoidance began with classical planning. Global methods
such as A$^\star$~\cite{hart1968} and
RRT$^\star$~\cite{karaman2011} compute near-optimal paths on a
known map, and reactive methods such as artificial potential
fields~\cite{khatib1986} and the dynamic window
approach~\cite{fox1997dwa} respond to sensed obstacles online. These
assume a base that executes commanded velocities faithfully, the
reactive ones suffer local minima and oscillation, and applied to a
biped they decouple avoidance from balance. CLF-CBF quadratic
programs~\cite{liu2023clfcbf} restore real-time safety guarantees
but still treat the gait as a separate tracking problem.

Reinforcement learning then removed the need for hand-designed
reactive rules, maturing first on wheeled robots for mapless
navigation~\cite{tai2017virtual}, goal-driven
exploration~\cite{cimurs2022}, and crowd
navigation~\cite{chen2017socially}. In parallel, model-free RL
produced robust legged locomotion, including periodic-reward gaits
on Cassie~\cite{siekmann2021}, transformer-based humanoid
locomotion~\cite{radosavovic2024science}, footstep-conditioned
walking~\cite{singh2022}, and massively parallel training for
quadrupeds~\cite{rudin2021,miki2022}. In our own prior work we
trained multi-skill locomotion for the same underactuated platform
using waypoint-based reward shaping~\cite{sahoo2025icc}. The two
lines rarely meet: navigation policies assume a statically stable
base, locomotion policies do not avoid obstacles, and the one
flat-RL attempt at bipedal avoidance~\cite{hamze2024} is limited to
fixed, known obstacle positions.

Hierarchical RL bridges the two. Temporal abstraction via
options~\cite{sutton1999options} and goal-conditioned hierarchies
such as HIRO~\cite{nachum2018hiro} showed that a coarse-timescale
high-level policy simplifies long-horizon tasks.
DeepLoco~\cite{peng2017deeploco} learned a footstep-level planner
over a stepping controller, hierarchical commands have been used
for quadruped navigation~\cite{jain2019hierarchical}, and recent
bipedal social navigation adopts a related
decomposition~\cite{shamsah2025}. Our work targets an underactuated
biped without roll joints, combines raycasts with explicit
dynamic-obstacle states at the HL, trains both levels jointly, and
benchmarks the learned HL against classical planners driving the
identical learned gait, which prior HRL locomotion studies do not.

\section{Preliminaries and Problem Formulation}

\subsection{Two-timescale decision process}
The LL control problem is a discrete-time Markov decision process
$(\mathcal{S}_{\ell}, \mathcal{A}_{\ell}, p_{\ell}, r_{\ell},
\gamma_{\ell})$ at $f_c = 50$~Hz, with $\gamma_{\ell} = 0.99$
(Table~\ref{tab:hparams}). Because the HL policy acts every $H=10$
LL steps with the action held constant, the HL problem is a
semi-MDP~\cite{sutton1999options} whose reward accumulates the
per-step task reward over the macro-step,
\begin{equation}
r_{h}(s_k, \mathbf{u}_k) =
\sum_{t=kH}^{(k+1)H-1} \gamma_{\ell}^{\,t-kH}\, r(s_t, a_t),
\label{eq:smdp}
\end{equation}
where $s_k$ is the HL state at macro-step $k$, $\mathbf{u}_k$ the
held command, $s_t$ the per-step LL state, and
$a_t \sim \pi_{\ell}(\cdot \mid s_t, \mathbf{u}_k)$ the LL action,
with $r(s_t,a_t)$ the task reward of Sec.~\ref{sec:hlpolicy}.
Eq.~\eqref{eq:smdp} shrinks the effective navigation horizon by a
factor of $H$, stabilizing credit assignment for long-range goal
reaching.

\subsection{Soft Actor--Critic}
Both levels are trained with SAC~\cite{haarnoja2018sac}, maximizing
the entropy-regularized return
\begin{equation}
J(\pi)=\sum_{t}\mathbb{E}_{(s_t,a_t)\sim\rho_\pi}
\!\left[r(s_t,a_t)+\alpha\,\mathcal{H}\!\left(\pi(\cdot\mid
s_t)\right)\right],
\label{eq:sacobj}
\end{equation}
where $\rho_\pi$ is the induced state--action distribution,
$\mathcal{H}$ the policy entropy, and $\alpha$ the temperature,
auto-tuned at both levels from $1.0$ (Table~\ref{tab:hparams}).
Eq.~\eqref{eq:sacobj} aids exploration in the contact-rich LL space
and the HL navigation problem alike, and off-policy SAC with twin
critics is more sample-efficient than on-policy alternatives such
as PPO~\cite{schulman2017ppo}.

\section{Method}

\begin{figure}[t]
\centering
\includegraphics[width=1\columnwidth]{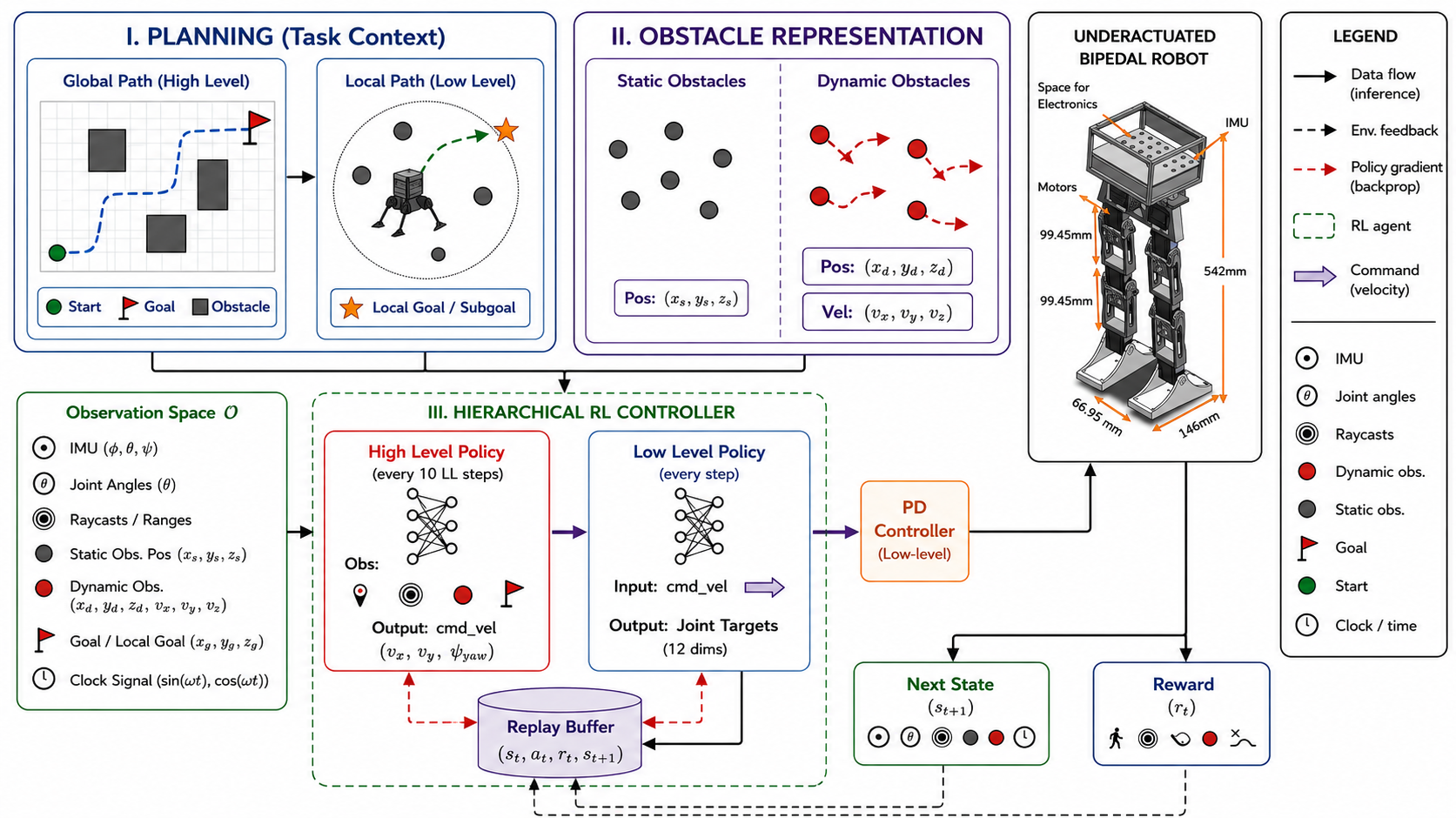}
\caption{System architecture: a global path supplies a
receding-horizon local goal; every $H{=}10$ steps the HL policy
outputs a velocity command executed by the LL policy through
PD-tracked joint targets; baselines replace only the HL block.}
\label{fig:arch}
\end{figure}

The control structure is shown in Fig.~\ref{fig:arch}.

\subsection{Planning layer and local-goal generation}
\label{sec:planlayer}
A global path is computed once per episode on the known static map.
Dynamic obstacles are excluded from the map, since handling them is
the HL policy's job. A receding-horizon module extracts a local
goal, the lookahead waypoint at arc-length $d_{la}$ (a tuned
constant) ahead of the robot's closest path point, advancing with
the robot and replaced by the final goal once within $d_{la}$. This
is the $\tilde g$ term of Eq.~\eqref{eq:hlobs}, and the same path
defines the tangent, cross-track error, and waypoint progress of
Eq.~\eqref{eq:rpath}. The layer issues no motion commands.

\subsection{Low-level velocity-conditioned gait policy}
\label{sec:llpolicy}
The LL observation
$s_{\ell} = [\,q, \dot q, \phi_{\text{base}}, \omega_{\text{base}},
c_t, \mathbf{u}\,]$ comprises the eight leg-joint positions $q$ and
velocities $\dot q$, base orientation $\phi_{\text{base}}$ and
angular velocity $\omega_{\text{base}}$ from the IMU, a clock signal
$c_t = [\sin\omega t, \cos\omega t]$, and the held command
$\mathbf{u}$; it has no obstacle or goal information.

The action is the target position for each of the eight actuated
joints, added to the half-sitting configuration and tracked by
PD controllers. With no hip or ankle roll, lateral commands $v_y$
are realized through coordinated hip-yaw and sagittal motion, so
the command box is asymmetric with
$|v_y| \le v_y^{\max} \ll v_x^{\max}$.

The LL reward rewards accurate tracking with a stable gait,
\begin{align}
r_{\ell} ={}& w_v\,
e^{-\lVert \mathbf{v}^{xy}_{\text{base}} - \mathbf{u}^{xy}
\rVert^2 / \sigma_v}
+ w_\omega\,
e^{-(\omega^{z}_{\text{base}} - \omega_{\text{yaw}})^2 /
\sigma_\omega}
\nonumber\\
&+ r_{\text{gait}},
\label{eq:llreward}
\end{align}
with measured planar velocity $\mathbf{v}^{xy}_{\text{base}}$ and
yaw rate $\omega^{z}_{\text{base}}$, empirically tuned weights
$w_v, w_\omega$ and tolerances $\sigma_v, \sigma_\omega$, and
$r_{\text{gait}}$ aggregating contact-timing
consistency~\cite{siekmann2021}, posture regularization,
action-rate smoothness, and torque penalties.

\subsection{High-level navigation policy}
\label{sec:hlpolicy}
The HL observation is
\begin{equation}
s_{h} = \big[\, p_{\text{base}}, \psi,\;
\mathbf{d}_{\text{ray}},\;
\{(\tilde p_j, \tilde v_j)\}_{j=1}^{N_d},\;
\tilde g \,\big],
\label{eq:hlobs}
\end{equation}
where $p_{\text{base}}, \psi$ are the base position and heading,
$\mathbf{d}_{\text{ray}} \in \mathbb{R}^{36}$ collects 36 raycasts
over $180^\circ$ clipped to $d_{\max}=3.0$~m, $\tilde p_j, \tilde
v_j$ are the position and velocity of the $j$-th of $N_d = 5$
nearest dynamic obstacles, and $\tilde g$ is the local goal in the
robot frame. Both raycasts and obstacle states are privileged
simulator quantities standing in for a LiDAR and object tracker
(Sec.~\ref{sec:limits}).

The HL action $\mathbf{u} = (v_x, v_y, \omega_{\text{yaw}})$ is
$\tanh$-squashed to the LL command box and held for $H=10$ steps,
so the HL policy runs at $5$~Hz. The task reward accumulates into
Eq.~\eqref{eq:smdp} as a shaping sum clipped at $-c_s$ ($c_s=15$)
plus terminal terms added after the clip,
\begin{equation}
r_t = \max\!\big(r_{\text{shape}},\, -c_s\big) + r_{\text{term}} .
\label{eq:hlreward}
\end{equation}
\begin{equation}
r_{\text{shape}} = r_{\text{path}} + r_{\text{posture}}
+ r_{\text{safe}} + r_{\text{reg}} .
\label{eq:rshape}
\end{equation}
The path term rewards progress along the reference,
\begin{equation}
r_{\text{path}} = w_{1}\Delta d_{wp} + w_{2} v_{\parallel}
+ w_{3}\big(\hat{\mathbf{h}}\!\cdot\!\hat{\mathbf{t}}\big)
- w_{4} e_{ct}^{2}
+ \frac{w_{9}}{d_{\text{goal}} + \epsilon},
\label{eq:rpath}
\end{equation}
with waypoint progress $\Delta d_{wp}$, path tangent
$\hat{\mathbf{t}}$, base heading $\hat{\mathbf{h}}$, forward
velocity $v_{\parallel}$, cross-track error $e_{ct}$, goal distance
$d_{\text{goal}}$, and $\epsilon = 10^{-5}$. The posture term makes
the HL accountable for gait feasibility,
\begin{align}
r_{\text{posture}} ={}& -w_{5}\big(|\phi_{r}| + |\phi_{p}|\big)
- w_{6} |z_b - z_{\text{ref}}|
- w_{7} v_{\perp}^{2} \nonumber\\
&- w_{8}\textstyle\sum_{i=1}^{8} |\tau_i|
+ w_{12}\!\!\textstyle\sum_{f \in \text{swing}}\!
\min(h_f, h_{\text{cap}}),
\label{eq:rgait}
\end{align}
with roll/pitch $\phi_r, \phi_p$, base height $z_b$
($z_{\text{ref}}=0.45$~m), lateral drift $v_\perp$, torques
$\tau_i$, and swing clearance $h_f$ capped at $h_{\text{cap}}$. The
safety term acts at two ranges,
\begin{equation}
r_{\text{safe}} =
- w_{14}\frac{\max\!\big(0,\, d_{\text{safe}} -
d_{\text{obs}}\big)}{d_{\text{safe}}}
- w_{15}\frac{\max\!\big(0,\, T_h - t_{\text{tc}}\big)}{T_h},
\label{eq:rsafe}
\end{equation}
with nearest-obstacle distance $d_{\text{obs}}$, safety radius
$d_{\text{safe}}=0.55$~m, time to collision (TTC) $t_{\text{tc}}$
from the nearest dynamic obstacle's relative state, and TTC horizon
$T_h=1.5$~s. The TTC term penalizes predicted collisions before
proximity alone would trigger, which is the key mechanism behind
dynamic-obstacle avoidance. The regularization term removes
degenerate optima,
\begin{equation}
r_{\text{reg}} = -w_{10}\,\mathbbm{1}_{\text{stuck}}
- w_{11}\,\mathbbm{1}_{\text{slow}} - w_{13},
\label{eq:rreg}
\end{equation}
where $\mathbbm{1}_{\text{stuck}}$ activates after
$t_{\text{stuck}}=0.7$~s without progress,
$\mathbbm{1}_{\text{slow}}$ when $v_{\parallel} <
v_{\min}=0.02$~m/s, and $w_{13}$ is a constant time penalty. Without
it, standing still near obstacles is favored. The terminal term is
\begin{equation}
r_{\text{term}} =
- R_{c}\,\mathbbm{1}_{\text{collision}}
- R_{f}\,\mathbbm{1}_{\text{fall}}
+ R_{g}\,\mathbbm{1}_{\text{goal}},
\label{eq:hlterm}
\end{equation}
with $R_c = R_f = 50$, $R_g = 30$. Collisions incur $-R_c$ without
terminating the episode, letting the policy learn recovery. Falls
and goal arrival terminate it. The clip $c_s$ prevents a transient
penalty pile-up from dominating the value function, and adding
Eq.~\eqref{eq:hlterm} after the clip keeps collision and fall
penalties undiluted. Weights $w_1$--$w_{15}$ were tuned manually
and are omitted for space. Deviating from the path to avoid an obstacle costs the cross-track term of Eq.~\eqref{eq:rpath} but is
recouped through Eq.~\eqref{eq:rsafe}. Resolving this trade-off is
what the HL policy learns.

\subsection{Joint two-timescale training}
\label{sec:training}
Both policies train simultaneously in the same episodes
(Algorithm~\ref{alg:training}). The LL policy is updated at 50~Hz
from Eq.~\eqref{eq:llreward}. The HL policy is updated at 5~Hz from
the macro-step return of Eqs.~\eqref{eq:smdp} and \eqref{eq:hlreward}.
Joint training couples the two levels: the gait adapts to the
navigation layer's actual command distribution, and the navigation
layer learns the gait's evolving feasibility limits through
Eq.~\eqref{eq:rgait}, with the HL policy acting as an automatic
curriculum. Episodes terminate on goal arrival, fall, or timeout.
After convergence the LL policy is frozen and reused by all
baselines of Sec.~\ref{sec:baselines}.

\begin{algorithm}[t]
\caption{Joint two-timescale HRL training}
\label{alg:training}
\begin{algorithmic}[1]
\STATE initialize $\pi_{h}$, $\pi_{\ell}$, buffers
$\mathcal{D}_h$, $\mathcal{D}_\ell$
\FOR{each episode}
  \STATE randomize layout/obstacles/start/goal; plan global path;
  init.\ local goal (Sec.~\ref{sec:planlayer})
  \FOR{each macro-step $k$}
    \STATE advance local goal; observe $s_h$
    (Eq.~\eqref{eq:hlobs}); sample $\mathbf{u}_k \sim \pi_h$
    \FOR{$t = kH$ \TO $(k{+}1)H - 1$}
      \STATE $a_t \sim \pi_{\ell}(\cdot \mid s_{\ell,t},
      \mathbf{u}_k)$; step environment; store in
      $\mathcal{D}_\ell$; update $\pi_\ell$
    \ENDFOR
    \STATE accumulate $r_h$; store in $\mathcal{D}_h$; update
    $\pi_h$
  \ENDFOR
\ENDFOR
\STATE freeze $\pi_\ell$ for reuse by all baselines
\end{algorithmic}
\end{algorithm}

\subsection{Planner-based baselines over the shared interface}
\label{sec:baselines}
All baselines drive the frozen co-trained LL policy, differing only
in how $\mathbf{u}_k$ is generated. In SAC+A$^\star$, the
planner~\cite{hart1968} plans on the inflated static map and a
pure-pursuit tracker converts the path to commands, with periodic
replanning from ground-truth obstacle positions in dynamic
scenarios as the strongest reasonable variant.
SAC+RRT$^\star$~\cite{karaman2011} plans on the same map and is
tracked identically. SAC+APF computes attractive and repulsive
forces from the same raycast and obstacle observations available
to our HL policy~\cite{khatib1986} and clips the resultant to the
command box. APF is fully reactive but susceptible to local minima.
Every method commands the same gait through the same interface at
the same rate, so Table~\ref{tab:baseline} compares navigation
intelligence only.

\section{Experiments}

All experiments are conducted in simulation.

\subsection{Setup}
The robot has four actuated joints per leg (hip yaw, hip pitch,
knee pitch, ankle pitch, eight DoF total), thigh and shank links of
$99.45$~mm, height $542$~mm, feet $146\times66.95$~mm, and a
base-mounted IMU. Simulation is in PyBullet (seed 42). Both levels
use $2\times256$ MLP actors and critics (Table~\ref{tab:hparams}),
trained on a single \texttt{NVIDIA RTX A4000} GPU.
Metrics are success, collision, and fall/timeout rate, path length,
and time to goal, over 10 unseen layouts per condition with 10
trials each (100 trials per method per condition).

\begin{table}[t]
\caption{SAC training hyperparameters, shared by both levels.}
\label{tab:hparams}
\centering
\small
\begin{tabular}{lc}
\toprule
Parameter & Value \\
\midrule
Discount factor $\gamma_{\ell}$      & $0.99$ \\
Learning rate                        & $1\times10^{-4}$ \\
Replay buffer size                   & $1\times10^{6}$ \\
Batch size                           & $256$ \\
Entropy temperature $\alpha$         & auto-tuned (init.\ $1.0$) \\
Soft update coefficient              & $0.005$ \\
Gradient steps per env.\ step        & $1$ \\
Max episode length                   & $5000$ steps ($100$~s) \\
Control frequency $f_c$              & $50$~Hz \\
HL decision period $H$               & $10$ steps ($5$~Hz) \\
Actor/critic networks                & MLP, $2\times256$ \\
Simulator                            & PyBullet (seed 42) \\
\bottomrule
\end{tabular}
\end{table}

\subsection{Static and dynamic scenarios}
\begin{figure}[t]
\centering
\includegraphics[width=1\columnwidth]{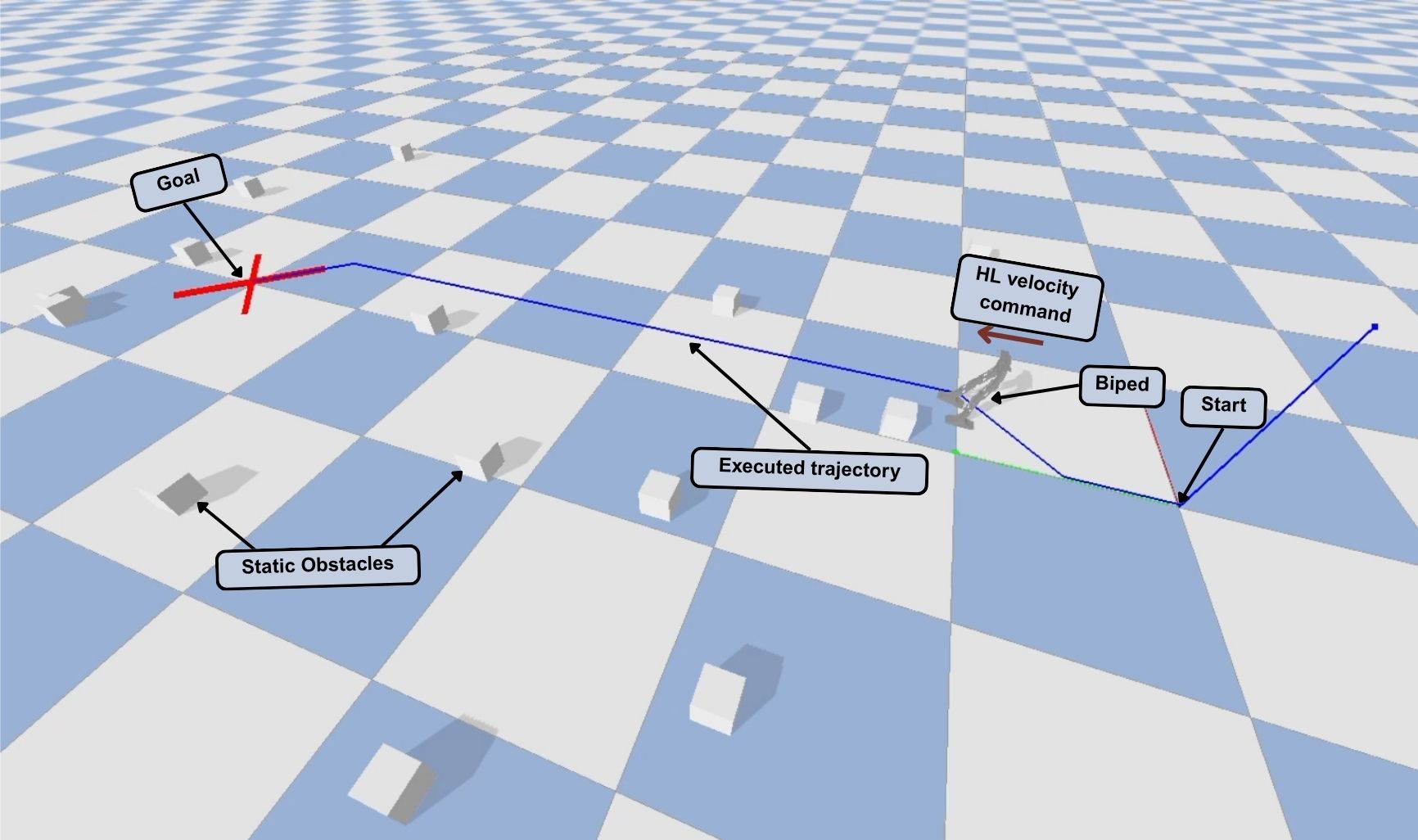}
\caption{Static-obstacle navigation scenario.}
\label{fig:static}
\end{figure}

\begin{figure}[t]
\centering
\includegraphics[width=1\columnwidth]{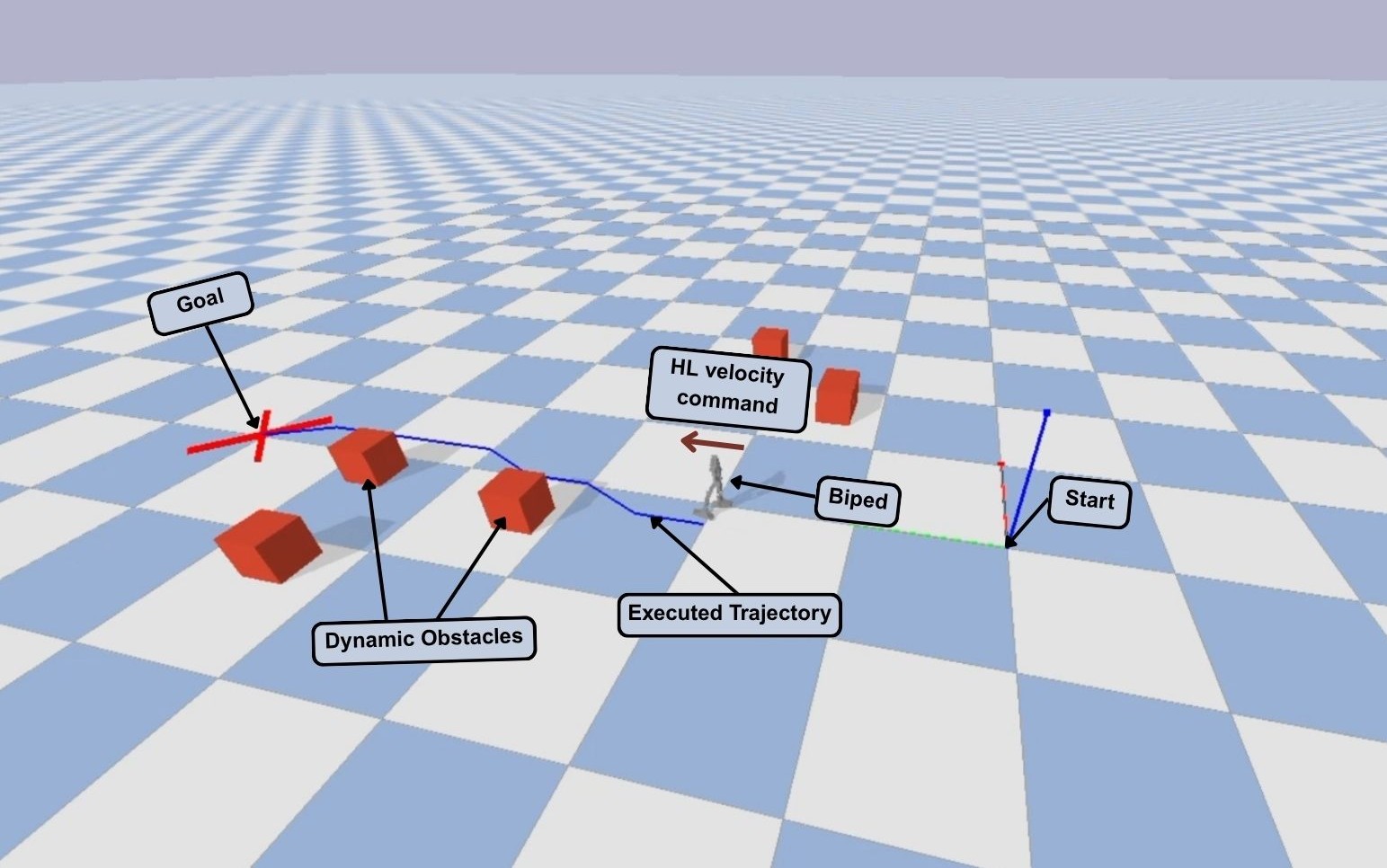}
\caption{Dynamic-obstacle navigation scenario.}
\label{fig:dynamic}
\end{figure}

Static evaluations use unseen randomized layouts with start and
goal $\ge 6.0$~m apart (Fig.~\ref{fig:static}). Dynamic obstacles
move along straight and lateral trajectories at $0.10$ to $0.30$~m/s
with up to $N_d = 5$ simultaneous movers, including crossing and
head-on encounters (Fig.~\ref{fig:dynamic}). The HL policy reduces
$v_x$ and issues a yaw-dominant command to let an obstacle pass,
then re-accelerates, while the LL policy preserves gait stability.

\begin{figure}[t]
\centering
\includegraphics[width=1\columnwidth]{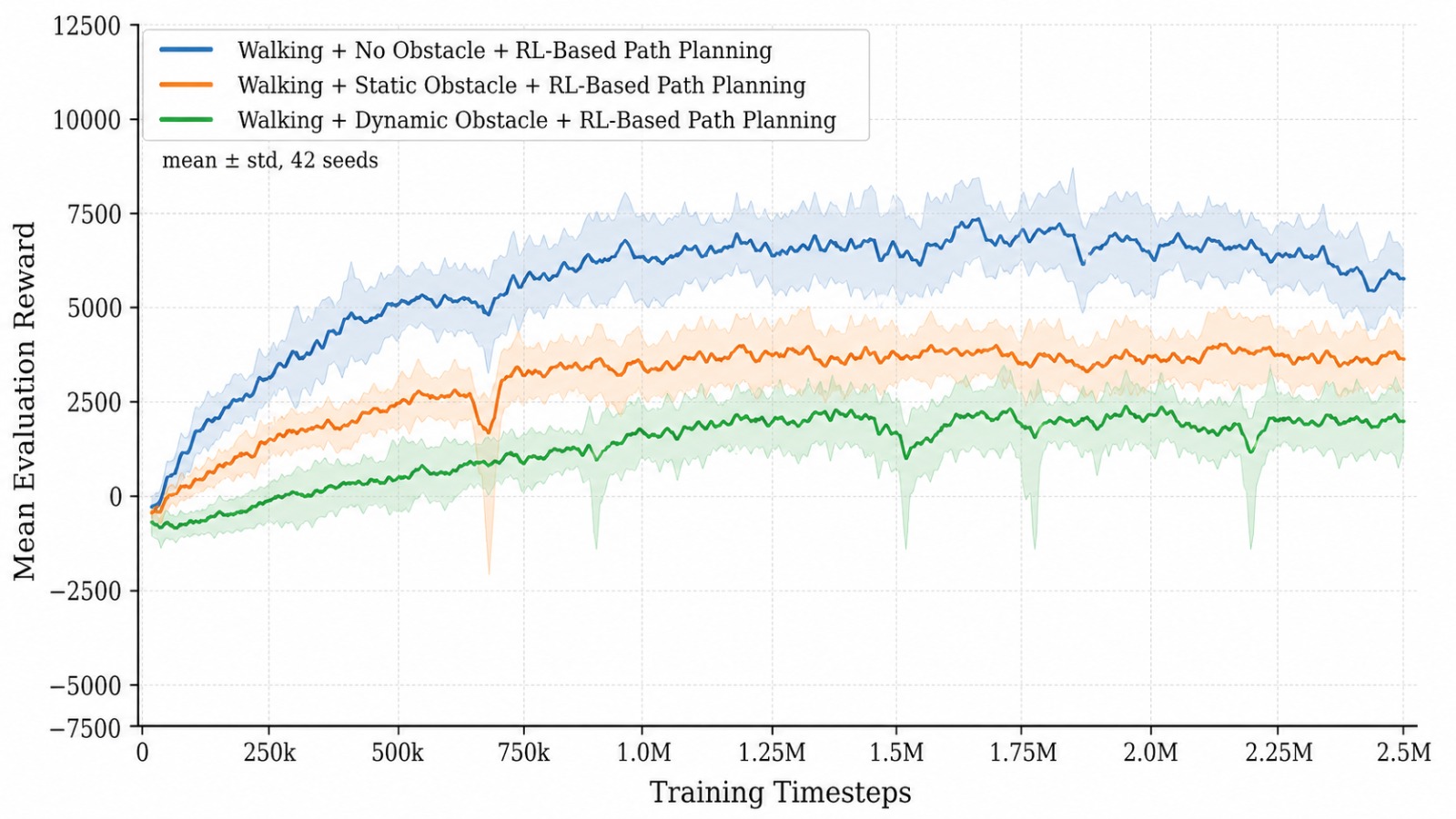}
\caption{Joint training curves under three conditions of increasing difficulty. Shaded bands denote per-evaluation variability.}
\label{fig:training}
\end{figure}

Fig.~\ref{fig:training} shows joint training curves under three
difficulty conditions. Each curve is the mean evaluation return
with a shaded band denoting per-evaluation variability of the
single run (seed 42). The dynamic-obstacle condition converges
more slowly since the policy must additionally learn velocity-aware
avoidance through the TTC term of Eq.~\eqref{eq:rsafe}.

\begin{table*}[!htbp]
\caption{Comparison with classical navigation baselines over 100
trials per method per condition. Path length and time are averaged
over successful episodes only.}
\label{tab:baseline}
\centering
\small
\begin{tabular}{lcccc|cccc}
\toprule
& \multicolumn{4}{c|}{Static} & \multicolumn{4}{c}{Dynamic} \\
Method & Succ.\ (\%) & Coll.\ (\%) & Path (m) & Time (s)
       & Succ.\ (\%) & Coll.\ (\%) & Path (m) & Time (s) \\
\midrule
SAC+A$^\star$~\cite{hart1968}      & 76.0 & 2.0  & 7.64 & 13.25 & 68.0 & 20.0 & 7.76 & 13.55 \\
SAC+RRT$^\star$~\cite{karaman2011} & 78.0 & 22.0 & 7.98 & 13.95 & 68.0 & 32.0 & 7.27 & 12.72 \\
SAC+APF~\cite{khatib1986}          & 63.0 & 37.0 & 6.87 & 14.27 & 55.0 & 45.0 & 6.80 & 13.12 \\
\textbf{HRL (ours)}                & \textbf{98.0} & \textbf{2.0} & 7.92 & 13.74 & \textbf{88.0} & \textbf{12.0} & 7.90 & 13.48 \\
\bottomrule
\end{tabular}
\end{table*}
\subsection{Comparison with planner-based hybrids}
Table~\ref{tab:baseline} compares the learned HL policy against the
three hybrids. On static maps the proposed method succeeds in
98.0\% of trials against a range of 76.0\% to 78.0\% for the global planners. In
dynamic scenarios the gap widens: the learned policy retains 88.0\%
success, whereas A$^\star$ and RRT$^\star$ collide in 20.0\% and
32.0\% of trials and APF drops to 55.0\% success with 45.0\%
collisions. Path lengths (7.92~m static, 7.90~m dynamic) remain
within 4\% of the A$^\star$ reference, so the safety gains are not
bought with detours. Per-episode averages are computed over
successful episodes only, so low-success methods report
optimistically short paths.

\subsection{Ablation study}
\begin{figure}[t]
\centering
\includegraphics[width=1\columnwidth]{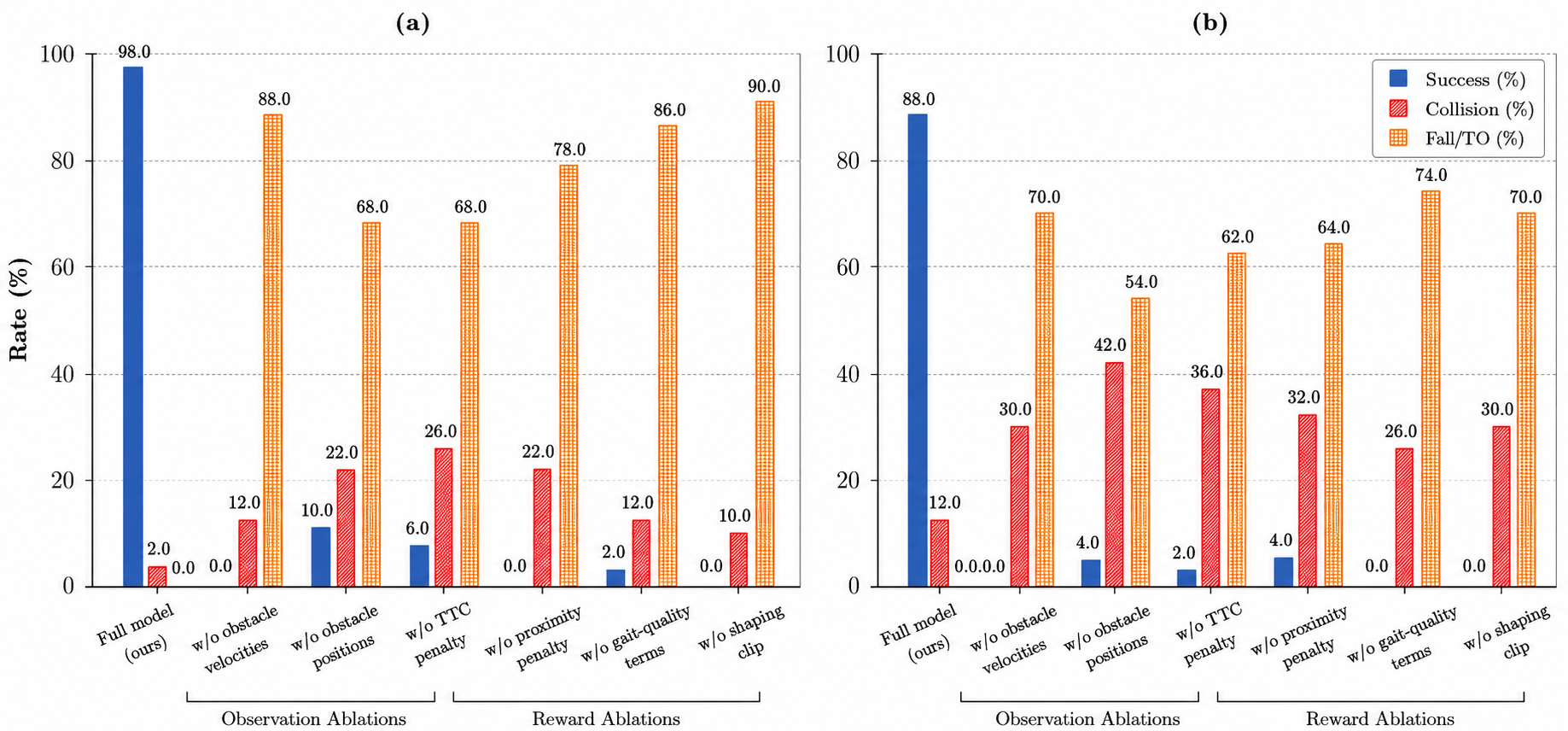}
\caption{Ablation of observation channels and reward terms in (a)
static and (b) dynamic environments.}
\label{fig:ablation}
\end{figure}

Fig.~\ref{fig:ablation} reports success, collision, and fall or
timeout rates for the full model and each ablated variant.
Removing the dynamic-obstacle velocity channel or obstacle
positions from Eq.~\eqref{eq:hlobs} degrades both regimes severely.
Removing the TTC or proximity penalty of Eq.~\eqref{eq:rsafe} has a
comparable effect, confirming anticipatory and proximal safety
signals are complementary. Removing the posture terms of
Eq.~\eqref{eq:rgait} shifts failures toward falls, and removing the
shaping clip of Eq.~\eqref{eq:hlreward} destabilizes training
outright (90.0\% fall/timeout, static). Across variants the
dominant failure mode is fall or timeout rather than collision,
indicating each ablation harms the navigation-balance coupling as a
whole.

\section{Discussion and Limitations}
\label{sec:limits}

The learned navigation layer's advantage rests on three mechanisms.
First, it commands only what the gait can execute: joint training
and Eq.~\eqref{eq:rgait} let it internalize the platform's
gait-limited, asymmetric command set, whereas the pure-pursuit
trackers of A$^\star$ and RRT$^\star$ demand infeasible turns,
surfacing as falls and timeouts even on static maps. Second, it
anticipates rather than reacts: the TTC term of
Eq.~\eqref{eq:rsafe} yields before proximity alone would trigger,
whereas global planners see moving obstacles only at replanning
instants, too late for a robot needing several steps to change
heading, explaining their dynamic collision rates of
20.0\% and 32.0\%, respectively. APF reacts instantly but on
instantaneous positions only, so it oscillates and stalls in local
minima, consistent with its 45.0\% dynamic collision rate.explaining their dynamic collision rates of
20.0\% and 32.0\%, respectively. APF reacts instantly but on
instantaneous positions only, so it oscillates and stalls in local
minima, consistent with its 45.0\% dynamic collision rate.
Third, the hierarchy assigns each problem to its natural timescale,
and the ablation, in which every removed component shifts failures
toward falls or timeouts rather than collisions, shows the
advantage comes from this coupled structure rather than any single
term.

Several limitations remain in this work. As this study is simulation-only. Sim-to-real
transfer would require addressing model mismatch, sensing noise,
and inter-level latency. The raycasts and dynamic-obstacle states
of Eq.~\eqref{eq:hlobs} are privileged simulator quantities standing
in for a LiDAR and object tracker, chosen to make that substitution
direct, though onboard perception remains future work. The planning
layer assumes a known static map, shared symmetrically by our
method and the baselines. Because the LL gait is co-trained with
the HL policy, the frozen gait reused by baselines may track their
commands slightly less faithfully. Results use a single training
seed, so seed variance is unquantified, and the policy actuates
only leg joints, leaving upper-body and multi-contact avoidance
open.

\section{Conclusion}

This paper presented a hierarchical reinforcement learning framework
for collision-free locomotion of an underactuated bipedal robot in
static and dynamic environments. A high-level SAC policy maps pose,
raycast, dynamic-obstacle, and local-goal observations to
body-velocity commands, and a velocity-conditioned low-level SAC
gait policy executes each command over ten control steps, with both
trained jointly at their respective timescales. The shared command
interface enabled a controlled comparison against SAC+A$^\star$,
SAC+RRT$^\star$, and SAC+APF hybrids driving the identical learned
gait, in which the proposed method attained 98.0\% success in
static and 88.0\% in dynamic scenarios within 4\% of the
shortest-path reference in path length. Ablations established that
the TTC term, proximity penalty, posture shaping, and reward clip
each contribute materially to this performance. Future work will
address onboard perception, seed-variance quantification, and
sim-to-real transfer.

\bibliographystyle{IEEEtran}
\bibliography{ref}

\end{document}